\documentclass[10pt,twocolumn,letterpaper]{article}

\usepackage{cvpr}
\usepackage{times}
\usepackage{epsfig}
\usepackage{graphicx}
\usepackage{amsmath}
\usepackage{amssymb}
\usepackage{booktabs}
\usepackage{url}
\usepackage{siunitx}
\usepackage{enumitem}

\usepackage[breaklinks=true,bookmarks=false]{hyperref}

\newcommand{\norm}[1]{\left\lVert #1 \right\rVert}
\newcommand{\abs}[1]{\left| #1 \right|}
\newcommand{\degr}{\si{\degree}}

\newcolumntype{C}[1]{>{\centering\let\newline\\\arraybackslash\hspace{0pt}}m{#1}}

\cvprfinalcopy 

\def\cvprPaperID{****} 

\begin{document}

\title{AdaLAM: Revisiting Handcrafted Outlier Detection}

\author{
Luca Cavalli\textsuperscript{1}, Viktor Larsson\textsuperscript{1}, Martin R. Oswald\textsuperscript{1}, Torsten Sattler\textsuperscript{2}, Marc Pollefeys\textsuperscript{1}\\
\textsuperscript{1} ETH Zurich, Switzerland\\
\textsuperscript{2} Chalmers University of Technology, Gothenburg, Sweden\\
{\tt\small lcavalli@ethz.ch}
}

\maketitle

\begin{abstract}
Local feature matching is a critical component of many computer vision pipelines, including among others Structure-from-Motion, SLAM, and Visual Localization. However, due to limitations in the descriptors, raw matches are often contaminated by a majority of outliers. As a result, outlier detection is a fundamental problem in computer vision, and a wide range of approaches have been proposed over the last decades.
In this paper we revisit handcrafted approaches to outlier filtering. Based on best practices, we propose a hierarchical pipeline for effective outlier detection as well as integrate novel ideas which in sum lead to AdaLAM, an efficient and competitive approach to outlier rejection. AdaLAM is designed to effectively exploit modern parallel hardware, resulting in a very fast, yet very accurate, outlier filter.
We validate AdaLAM on multiple large and diverse datasets, and we submit to the Image Matching Challenge (CVPR2020), obtaining competitive results with simple baseline descriptors.
We show that AdaLAM is more than competitive to current state of the art, both in terms of efficiency and effectiveness.
\end{abstract}

\section{Introduction} \label{sec:introduction}
%
Image matching is a key component in any image processing pipeline that needs to draw correspondences between images, such as Structure from Motion (SfM)~\cite{ullman1979interpretation,hartley1997triangulation,schoenberger2016sfm,schoenberger2016mvs,wu2011visualsfm,moulon2012adaptive}, Simultaneous Localization and Mapping (SLAM)~\cite{bailey2006simultaneous,durrant2006simultaneous,montemerlo2002fastslam} and Visual Localization~\cite{li2010location,cech2010efficient,sattler2018benchmarking,sarlin2019coarse}. Classically, the problem is tackled by computing high dimensional descriptors for keypoints which are robust to a set of transformations, then a keypoint is matched with its most similar counterpart in the other image, \ie the nearest neighbor in descriptor space. Due to limitations in the descriptors, the set of nearest neighbor matches usually contains a great majority of outliers as many features in one image often have no corresponding feature in the other image. 
Consequently, outlier detection and filtering is an important problem in these applications. Several methods have been proposed for this task, from simple low-level filters based only on descriptors such as the ratio-test~\cite{lowe2004distinctive}, to local spatial consistency checks~\cite{sattler2009scramsac,bian2017gms,ma2019locality,sivic2008efficient,cech2010efficient,torr2002napsac,ni2009groupsac,jung2001robust,zhang1995robust,lin2017code,wu2009bundling,wu2015robust,albarelli2010robust,johns2015ransac,leordeanu2005spectral} and global geometric verification methods, either exact~\cite{fischler1981random,torr2002napsac,chum2005matching,ni2009groupsac,johns2015ransac,chum2008optimal,torr2000mlesac,chum2003locally,lebeda2012fixing,barath2018graph,raguram2012usac,barath2019magsac} or approximate~\cite{jegou2008hamming,avrithis2014hough,li2015pairwise,wu2015adaptive,wu2015robust,schonberger2016vote}. In the last two decades, many methods have been proposed to learn either local neighborhood consistency~\cite{rocco2018neighbourhood,zhao2019nm} or global geometric verification~\cite{zhang2019oanet,sarlin2019superglue,brachmann2019neural,ranftl2018deep,moo2018learning,dang2018eigendecomposition}.
\begin{figure*}[t]
    \centering
    \includegraphics[width=0.8\textwidth]{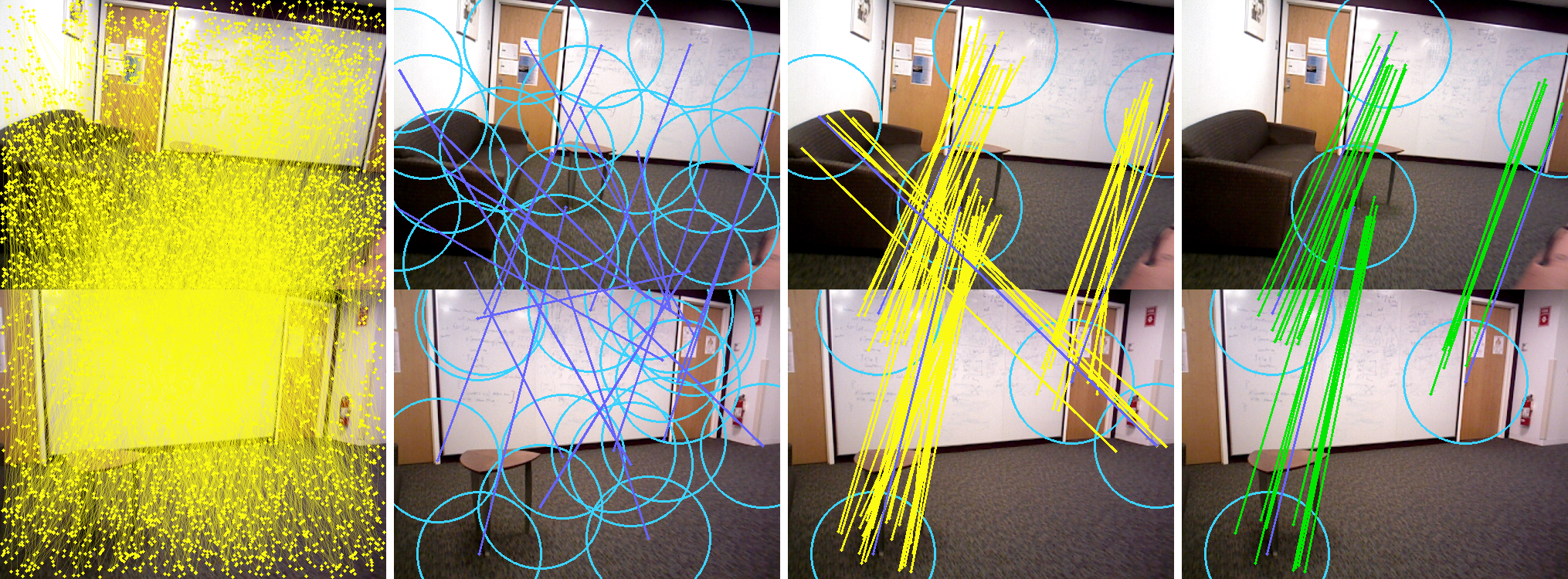}
    \caption{\textbf{Main steps in AdaLAM}, from left to right: \textit{1.} we take as input a wide set of putative matches (in yellow), \textit{2.} we select well spread hypotheses of rough region correspondences (blue circles), \textit{3.} for each region we consider the set of all putative matches consistent with the same region correspondence hypothesis, \textit{4.} we only keep the correspondences which are locally consistent with an affine transform with sufficient support (in green). Note that for visualization purposes we do \textit{not} show all the hypotheses \textit{nor} all the matches.}
    \label{fig:schematic}
\end{figure*}

In this paper we revisit handcrafted approaches to outlier filtering. Following best practices from this mature field of research, we propose a hierarchical pipeline for efficient and effective outlier filtering based on local affine motion verification with sample-adaptive threshold. We name our approach Adaptive Locally-Affine Matching (AdaLAM). We show that AdaLAM achieves more than competitive performance to current state of the art in both outdoor and indoor scenes. Moreover, we design AdaLAM to effectively exploit modern parallel hardware, taking less than 20 milliseconds per image pair for producing filtered matches from 8000 keypoints per image on a modern GPU.

We can summarize our contributions in the following:
\begin{itemize}[itemsep=1pt,topsep=0pt,leftmargin=*]
    \item We propose AdaLAM, a novel outlier filter that builds up from several past ideas in spatial matching into a coherent, robust, and highly parallel algorithm for fast spatial verification of image correspondences.
    \item As our framework is based on geometrical assumptions that can have different discriminative power in different scenarios, we propose a novel method that adaptively relaxes our assumptions, to achieve better generalization to different domains while still mining as much information as available from each image region.
    \item We experimentally show that our adaptive relaxation improves generalization, and that AdaLAM can greatly outperform current state-of-the-art methods.
\end{itemize}




\section{Related Work}
Outlier rejection is a long-standing problem which has been studied in many contexts, producing many diverse approaches that act at different levels, with different complexity and different objectives.

\noindent
\textbf{Simple filters} are widely used as a straightforward heuristic that already greatly improves the inlier ratio of available correspondences based on very low-level descriptor checks. In this category we include the classical ratio-test~\cite{lowe2004distinctive} and mutual nearest neighbor check, that filter out ambiguous matches, as well as hamming distance thresholding to prune obvious outliers. These heuristics are extremely efficient and easy to implement, though they are not always sufficient as they can easily leave many outliers or filter out inliers present in the initial putative matches set.

\noindent
\textbf{Local neighborhoods} methods filter correspondences based on the observation that correct matches should be consistent with other correct matches in their vicinity, while wrong matches are normally inconsistent with their neighbors. Consistency can be formulated as a co-neighboring constraint~\cite{sattler2009scramsac,bian2017gms,ma2019locality,sivic2008efficient,cech2010efficient,torr2002napsac,ni2009groupsac}, or enforcing a local transformation between neighboring correspondences~\cite{jung2001robust,zhang1995robust,lin2017code,wu2009bundling,wu2015robust}, or as a graph of mutual pairwise agreements of local transformations~\cite{albarelli2010robust,johns2015ransac,leordeanu2005spectral}.
Methods acting at this level can also be very efficient, and represent a more informative selection compared to simple filters.

\noindent
\textbf{Geometric verification} approaches filter matches based on a global transformation on which correct correspondences must agree. This can be achieved by robustly fitting a global transformation (be it similarity, affinity, homography or fundamental) to the set of all the matches, with sampling methods, including RANSAC~\cite{fischler1981random} and its numerous later improvements, either biasing the sampling probabilities towards more likely inliers~\cite{torr2002napsac,chum2005matching,ni2009groupsac,johns2015ransac}, making iterations more efficient with a sequential probability ratio test~\cite{chum2008optimal} or adding local optimization~\cite{torr2000mlesac,chum2003locally,lebeda2012fixing,barath2018graph}, combining all of the previous~\cite{raguram2012usac}, or marginalizing over the inlier decision threshold~\cite{barath2019magsac}. There is also a line of research dealing with the problem of estimating the inlier noise level~\cite{cohen2015likelihood,konouchine2005amlesac} or elaborating alternative noise-independent metrics for inlier selection~\cite{moisan2004probabilistic}.
A different line of research in the context of image retrieval uses fast approximate spatial verification to determine whether two images have the same content. They only approximately fit a geometric transformation to efficiently prune the majority of outliers, using the local affine or similarity transformation encoded by single correspondences~\cite{lowe2004distinctive}. The space of all transforms is quantized and the set of accepted correspondences is determined by majority voting with a hough scheme in linear time~\cite{jegou2008hamming,avrithis2014hough,li2015pairwise,wu2015adaptive,wu2015robust,schonberger2016vote}.

\noindent
\textbf{Learned methods} extract an implicit consistency model directly from data. Several works have been proposed in the last years, acting on different levels, either learning a local neighborhood consistency model~\cite{rocco2018neighbourhood,zhao2019nm}, or a global consistency model~\cite{zhang2019oanet,sarlin2019superglue,brachmann2019neural,ranftl2018deep,moo2018learning,dang2018eigendecomposition}. Many works target learning epipolar geometry constraints explicitly, formulating the problem either as outlier classification~\cite{moo2018learning,dang2018eigendecomposition}, or as an iteratively reweighted least-squares problem~\cite{ranftl2018deep}, or biasing RANSAC sampling distribution~\cite{brachmann2019neural}.

In this paper, we integrate best practices from the vast literature of handcrafted methods matured in this field into a coherent framework for fast and effective outlier filtering. In our ablation studies we show that a well-engineered baseline built on known good practices already achieves comparable performance to current state of the art in specific settings. We analyse the limitations of such baseline and propose a novel adaptive thresholding scheme to improve and generalize performance to a wide range of diverse settings.

\section{Hierarchical Adaptive Affine Verification} \label{sec:method}
%
Given the sets of keypoints $\mathcal{K}_1$ and $\mathcal{K}_2$ respectively in images $I_1$ and $I_2$, in this work we consider the set of all putative matches $\mathcal{M}$ to be the set of nearest neighbor matches from $\mathcal{K}_1$ to $\mathcal{K}_2$. In practice, due to limitations in the descriptors, $\mathcal{M}$ is contaminated by a great majority of incorrect correspondences, thus our objective is to produce a subset $\mathcal{M}' \subseteq \mathcal{M}$ that is the nearest possible approximation of the set of all and only correct inlier matches $\mathcal{M}^* \subseteq \mathcal{M}$.

Our method builds on classical spatial matching approaches used both in the field of matching and image retrieval. To keep computational costs down, we limit our search of matches to the set of initial putative matches $\mathcal{M}$ of the nearest neighbors in descriptor space. The main steps in our algorithm are reported in Figure~\ref{fig:schematic} and can be summarized as follows:
\begin{enumerate}[itemsep=1pt,topsep=0pt,leftmargin=*]
    \item We select a limited number of confident and well distributed matches, which we call seed points.
    \item For each seed point we select neighboring compatible correspondences.
    \item We verify local affine consistency in neighborhoods of each seed point by running highly parallel RANSACs~\cite{fischler1981random} with sample-adaptive inlier thresholds. 
    We output $\mathcal{M}'$ as the union of all the set of inliers of the seed points with strong enough support within each one's specific inlier threshold.
\end{enumerate}



\subsection{Preliminaries and core assumptions} \label{subsec:assumptions}
The 3D plane tangent to a point induces an homography between two views, which can be well approximated locally by an affine transformation $A$ in image space~\cite{koser2009geometric}.
This affine transformation strongly constraints geometrical cross consistency of correct keypoint correspondences, acting as a very reliable filter. However, the underlying assumptions of planarity, locality and correct projections can break in multiple ways in real images:
\begin{enumerate}[itemsep=1pt,topsep=0pt,leftmargin=*]
    \item The surface on which 3D points lie may not be planar. The offset between the 3D tangent plane at a point and the real surface produces a non-linear deviation in the projections of all the 3D points not lying on the tangent plane, which is more and more significant with the curvature of the surface.
    \item The detected points may not be near to each other, adding distortion to the affine model which is no longer a good approximation of the induced homography. This error increases with the relative distance of keypoints and with the tilt of the tangent plane.
    \item Matching keypoints may not represent the projection of exactly the same 3D point. This is a very common problem with wide baseline viewpoint changes, as slight changes in illumination and self occlusions can easily move the peak in saliency for keypoint localization.
    \item Incorrect non-linear lens distortion models can further introduce non-linearities in the motion of points between two views.
\end{enumerate}
To address these problems we propose an adaptive relaxation on our core assumption, that we describe in Section \ref{subsec:adasoft}

\subsection{Seed points selection}
As affine transforms $A$ are a good approximation of local transformations around a 3D point $P$, we use available nearest neighbor correspondences to guide the search for candidate 3D surface points. More specifically we want to select a restricted set of confident and well spread correspondences to be used as hypotheses for $P$, around which consistent point correspondences are to be searched, as in~\cite{jung2001robust}. We call such hypotheses \textit{seed points}. For the selection of seed points, we assign a confidence score to each keypoint and then promote a keypoint to seed point if it has the highest score within a radius $R$. This is implemented efficiently as a local non-maximum suppression over the scores. As a confidence score, we use the well known ratio-test of the nearest neighbor correspondence assigned to a keypoint. This way we ensure both distinctiveness and coverage of seed points without causing grid artifacts, while keeping the selection completely parallel for efficient computation on GPU, as each correspondence can be scored and compared to neighbors for seed point selection independently of the final selection of the others.

\subsection{Local neighborhood selection and filtering}
The assignment of correspondences to seed points is a crucial step in the algorithm as it builds the search space around each hypothesis of $P$ to find the affine transform $A$. Wider neighborhoods can more easily include correct correspondences to fit $A$, while at the same time they implicitly loosen the affine constraints as they violate the assumption on locality.

Let $S_i = (\mathbf{x}_1^{S_i}, \mathbf{x}_2^{S_i})$ be a seed point correspondence, which induces a similarity transformation $(\alpha^{S_i} = \alpha_2^{S_i}-\alpha_1^{S_i}, \sigma^{S_i}=\sigma_2^{S_i}/\sigma_1^{S_i})$ from its local feature frame, decomposed in the orientation component $\alpha^{S_i}$ and scale component $\sigma^{S_i}$, and $\mathcal{N}_i \subseteq \mathcal{M}$ be the set of correspondences that are assigned to $S_i$ to verify affine consistence. Let $t_\alpha$ and $t_\sigma$ be thresholds for orientation and scale agreement between a candidate correspondence and the seed correspondence $S_i$. In analogy to~\cite{sattler2009scramsac}, correspondence $(p_1, p_2) = ((\mathbf{x}_1, \mathbf{d}_1, \sigma_1, \alpha_1), (\mathbf{x}_2, \mathbf{d}_2, \sigma_2, \alpha_2)) \in \mathcal{M} $, which induces a transformation $(\alpha^p = \alpha_2-\alpha_1, \sigma^p=\sigma_2/\sigma_1)$ is assigned to $\mathcal{N}_i$ if the following constraints are satisfied:
\begin{gather}
    \norm{\mathbf{x}_1^{S_i}\!-\!\mathbf{x}_1} \leq \lambda R_1 \wedge \norm{\mathbf{x}_2^{S_i}\!-\!\mathbf{x}_2} \leq \lambda R_2 \label{eq:corrinclusiondistance} \\ 
    \abs{\alpha^{S_i}\!-\! \alpha^p} \leq t_\alpha \wedge \abs{\text{ln}\!\!\left(\frac{\sigma^{S_i}}{\sigma^p}\right)} \leq t_\sigma \label{eq:corrinclusionsideinfo}
\end{gather}
where $R_1$ and $R_2$ are the radii used to spread seed points respectively in image $I_1$ and $I_2$, and $\lambda$ is a hyperparameter that regulates the overlap between inclusion neighborhoods. Note that we consider angles $\alpha$ in modulo $2\pi$ lying within the interval $(-\pi, \pi]$. Different radii $R_1$ and $R_2$ are chosen proportionally to the image area to be invariant to image rescaling.

As from Eq.~\eqref{eq:corrinclusiondistance}, we include in $\mathcal{N}_i$ all the correspondences in $\mathcal{M}$ that are locally consistent in both images within a radius of $\lambda R$, \ie all the correspondences that match roughly in the same regions as the seed matches. According to Eq.~\eqref{eq:corrinclusionsideinfo}, we further filter sets $\mathcal{N}_i$ to ensure that included correspondences induce a similarity transform $(\alpha^p, \sigma^p)$ which is consistent with $(\alpha^{S_i}, \sigma^{S_i})$ within independent thresholds $t_\alpha$ and $t_\sigma$. The independent thresholds encode a confidence over the reliability of the orientation and scale information provided by keypoints. The idea of verification using orientation and scale consistency has been repeatedly proposed for template matching~\cite{lowe2004distinctive,albarelli2010robust} and image retrieval~\cite{schonberger2016vote,avrithis2014hough,jegou2008hamming} as a rough but powerful indication for outlier pruning.

At this stage, each set $\mathcal{N}_i$ can be processed independently for affine verification.

\subsection{Adaptive Affine Verification} \label{subsec:adasoft}
In this section we describe how we select inliers within a neighboring set $\mathcal{N}_i$. We approach the problem with a classical RANSAC framework~\cite{fischler1981random}, running a fixed number of iterations with minimal samples. Notice that a minimal sample for a centered affine transformation consists of only two correspondences, thus we do not need to run many iterations in practice. The sampling scheme is directly inspired from PROSAC~\cite{chum2005matching}. We use the ratio-test score to bias the RANSAC sampling distribution to give precedence to samples which are more likely to be inliers. However, PROSAC is looking for the best model to explain data with the guarantee that such a model exists, while in our setting we do not want to spend excessive computational power on neighboring sets $\mathcal{N}_i$ that may be outliers themselves. For this reason, we only run a fixed number of iterations for all neighboring sets $\mathcal{N}_i$, in contrast with PROSAC's sampling strategy which gradually degenerates to uniform sampling as long as early termination is not triggered. Moreover, we deterministically select the samples to be drawn so that we exhaustively explore the most likely minimal samples as much as the fixed iterations budget allows.

At each iteration $j$ we sample a minimal set and fit the affine transformation $A_i^j$ for each neighboring set $\mathcal{N}_i$. A residual $r_k$ is assigned to correspondence $k$ with points $\mathbf{x}^k_1, \mathbf{x}^k_2$ to measure its deviation from $A_i^j$:
\begin{equation}
     r_k(A_i^j) = \norm{A_i^j\mathbf{x}^k_1 - \mathbf{x}^k_2}
\end{equation}
As anticipated in Section~\ref{subsec:assumptions}, the affine model only partially explains the image motion that we want to recognize, and there is no clear bound to the error of the affine model. As a consequence, we cannot set a fixed threshold over $r_k$ to determine whether $k$ is an inlier. We take inspiration from~\cite{moisan2004probabilistic}: instead of thresholding directly on the error score, we threshold on the statistical significance of an inlier set against the null hypothesis of uniformly scattered outliers. This mapping better translates our objective: we do not bound the deviation of the affine model from the real motion model, but rather the likelihood of the observations.

In particular, if $\mathcal{H}_o$ is the hypothesis of having uniformly scattered outlier correspondences, we map a residual value $r_k$ within a set of residuals $\mathcal{R}$ to a confidence measure $c_k$:
\begin{equation}
    c_k(\mathcal{R}) = \frac{P}{\mathbb{E}_{\mathcal{H}_o}[P]} = \frac{P}{|\mathcal{R}|\frac{r_k^2}{R_2^2}}
\end{equation}
where the positive samples count $P = |{l: r_l \leq r_k}|$ is the number of inliers assuming that correspondence $k$ is the worst inlier. Notice that sorting $\mathcal{R}$ by residual yields $P=k+1$ (assuming 0-indexing). The proposed confidence $c$ effectively measures the ratio between the number of inliers actually found and the number of inliers that would be found under the outlier-only hypothesis $\mathcal{H}_o$. It can be interpreted as a ratio-test that compares the outlier-only hypothesis with the sample evidence. Our substitution $\mathbb{E}_{\mathcal{H}_o}[P] = |\mathcal{R}|\frac{t^2}{R_2^2}$ is done under the assumption that outliers are distributed uniformly over the whole sampling radius $R_2$ in the second image. Even when the uniformity assumption fails, the proposed metric deviates linearly with the increased local density of outliers. In practice this happens very often as correspondences are clustered in highly textured regions which only cover a small portion of the sampling radius. However, using conservatively high thresholds $t_c$ on the value of $c$, significant patterns in data can still emerge with very high confidence.

In complete analogy with the classical thresholding on the geometric error, we mark as inliers the samples $k$ such that $c_k \geq t_c$. While this implements a significance-driven adaptive thresholding scheme, still our inverse confidence is extremely simple and fast to compute on parallel hardware.

We compute the confidence of all samples for each iteration in parallel, and select inliers with a fixed threshold $t_c$ on the confidence. The affine model is fit again with the set of inliers, and inliers are selected again according to the new model. Finally, for each set $\mathcal{N}_i$ we only output the inliers of the iteration that scored the highest inlier count, and further filter out the $i$ scoring less than $t_n$ inliers for their best iteration.

\subsection{Implementation details}
As modern learning approaches run efficiently on highly parallel hardware, we also designed our algorithm to be extremely parallel to run efficiently on modern GPUs, and accordingly we provide a full implementation in PyTorch~\cite{paszke2017automatic}. This allows a great speedup compared to CPU execution, although it still leaves a wide space for further low-level optimization.

In particular seed point selection is implemented as a local non-maximum suppression, where each correspondence is evaluated independently of the evaluation of the others. For affine fitting, random sampling methods such as RANSAC suit perfectly the needs as all different seed points as well as all iterations can be processed in parallel.

For the following experiments, we set the radius $R$ for seed point selection to match a fixed ratio $r_a$ between the area of the non-maximum suppression circle $R^2\pi$ and the area of the image $wh$. In particular, we set $R = \sqrt{\frac{wh}{\pi r_a}}$ with $r_a=100$. For the purpose of collecting neighborhoods $\mathcal{N}_i$ for each seed point $i$, we use a radius $\lambda$ times larger than $R$, with $\lambda =4$. This ensures sufficient but controlled overlap between neighboring regions to be robust to errors in seed correspondences. We observed that the performance of our method saturates very early when increasing the number of RANSAC iterations, which are fixed to 128 iterations. When experimenting with SIFT keypoints, we set $t_\sigma = 1.5$ and $t_\alpha = 30\degr$. Finally, we observed that best performance is obtained with very conservative values of $t_c$, thus we set $t_c=200$ and require a verified neighboring set to have at least 6 inliers to be accepted. 

\section{Experiments}
Our experiments aim at comparing our method with existing state-of-the-art methods, and to understand the influence of each proposed component with ablation studies. The results of these experiments are reported respectively in Section~\ref{subsec:compsota} and~\ref{subsec:ablation}. We measure relative pose estimation performance under the same pipeline and on the same datasets for both comparative and ablation studies. We evaluate on the same test sets as OA-Net~\cite{zhang2019oanet}, NG-RANSAC~\cite{brachmann2019neural} and GMS~\cite{bian2017gms}, which we compare to: the same four scenes from YFCC100M~\cite{thomee2016yfcc100m}, two from Strecha~\cite{strecha2008benchmarking} and fifteen from SUN3D~\cite{xiao2013sun3d} as~\cite{zhang2019oanet,brachmann2019neural}, and the same six sequences from TUM~\cite{sturm2012benchmark} as~\cite{bian2017gms}.
While affine verification is naturally competitive in man-made outdoor scenarios with many well textured and planar scenes, the ablation studies show that our adaptive thresholding is particularly effective in generalizing such advantage to more challenging indoor scenes. As a result, we show that AdaLAM can greatly outperform the current state of the art in both indoor and outdoor scenarios. 

Moreover, we submit AdaLAM to compete in two publicly available challenges. We discuss submission details and results for the submission to the Aachen Day-Night Challenge~\cite{sattler2018benchmarking,sattler2012image} in Section~\ref{subsec:aachendaynight}, and for the Image Matching Challenge (CVPR 2020)~\cite{jin2020image} in Section~\ref{subsec:imagematchingchallenge}

\subsection{Evaluation Pipeline}
Our evaluation pipeline aims at measuring relative pose estimation performance within the same settings. More specifically, all methods receive exactly the same keypoints in input and need to output a set of matches that will be used to robustly fit an essential matrix, which is decomposed to rotation and translation. We then measure the rotation and translation errors in degrees and take the maximum of the two, and report the exact Area Under the Curve (AUC) with thresholds of 5, 10, and 20 degrees.

The keypoints are all extracted with OpenCV SIFT~\cite{lowe2004distinctive} with the same parameters as in the code provided by OA-Net~\cite{zhang2019oanet} and NGRANSAC~\cite{brachmann2019neural}, with a maximum number of 8000 keypoints per image. Keypoints with locations, descriptors, orientation and scale are provided to the matching methods, and matches are produced. For fitting the essential matrix we use the LO-RANSAC~\cite{chum2003locally} implementation in COLMAP~\cite{schoenberger2016sfm,schoenberger2016mvs} with minimum $10^3$ iterations and maximum $10^4$. The calibration is assumed to be known and is taken from ground truth.

\begin{table}[tb]
\scriptsize
\centering
\caption{\textbf{Comparative experiments} with the state of the art in indoor and outdoor scenes. All methods fit the essential matrix with LO-RANSAC with maximum $10^4$ iterations, except \textit{Ratio test (100k)} that uses $10^5$. All AUC values are in percentages. We also report the F1-score of all method outputs before LO-RANSAC relative to the ground truth inliers. \textit{All methods have been tuned for AUC}.}
\label{tab:comparisons}
\setlength{\tabcolsep}{1.0mm}
\begin{tabular}{lccccccccc}
\toprule
 Method & 
 \multicolumn{3}{c}{TUM~\cite{sturm2012benchmark}} & 
 \multicolumn{3}{c}{SUN3D~\cite{xiao2013sun3d}} & 
 \multicolumn{3}{c}{YFCC100M~\cite{thomee2016yfcc100m}}\\
\cmidrule(r){2-4}
\cmidrule(rl){5-7}
\cmidrule(r){8-10}

%
 & AUC5 & AUC20 & F & AUC5 & AUC20 & F & AUC5 & AUC20 & F\\
\midrule

%
AdaLAM                                       & \bf 27.4& \bf 53.4 & \bf 49.1 & \bf 7.9 & \bf 48.4 & \bf 34.6 &\bf 58.8& \bf 82.4 & 59.4 \\
OA-Net~\cite{zhang2019oanet}                 & 22.1& 43.8& 48.3 & 7.0 & 29.7 & 47.8 & 54.8& 77.5& \bf 66.6 \\
GMS~\cite{bian2017gms}                       & 19.6& 41.3& 23.9 & 6.8 & 29.1 & 24.5 & 52.3& 76.0& 40.0 \\
NG-RANSAC~\cite{brachmann2019neural}         & 19.4& 38.7& 27.3 & 6.2 & 27.3& 30.5  & 53.8& 77.7& 45.5 \\
Ratio test (10k)  & 16.1& 33.6& 24.7 & 5.9 & 25.6 & 26.9 & 51.9& 76.3& 42.1 \\
Ratio test (100k) & 17.3& 36.2& 24.7 & 6.1 & 26.3 & 26.9 & 53.2& 77.5& 42.1 \\
\bottomrule
\end{tabular}
\end{table}

\subsection{Datasets}
We evaluate our method on large and diverse indoor and outdoor datasets, using the same scenes as the methods we compare with.
    For outdoor scenes we use the YFCC100M~\cite{thomee2016yfcc100m} internet photos, that were later organized into 72 scenes~\cite{heinly2015_reconstructing_the_world} reconstructed with the Structure from Motion software VisualSfM~\cite{wu2011visualsfm,wu2011multicore}, providing bundle adjusted camera poses, intrinsics and triangulated point clouds. We select scenes and image couples as to reproduce the test set used by~\cite{brachmann2019neural,moo2018learning,zhang2019oanet}, thus we used the same six scenes, including the two from Strecha~\cite{strecha2008benchmarking}, with the same sampling procedure and using the original image resolution. From now on when we refer to YFCC100M, we are referring to the four scenes actually coming from YFCC100M \textit{and} the two coming from Strecha.
    
    For indoor scenes we use six sequences from the TUM~\cite{sturm2012benchmark} visual odometry benchmark and the SUN3D~\cite{xiao2013sun3d} dataset, both of which provide ground truth poses together with the RGB images. In particular, for TUM we select the same sequences as the authors of GMS~\cite{bian2017gms}, but we use a different subsampling scheme to provide a wider range of image transformations. We take one keyframe every 150 frames, and match it with other 9 images sampled at 15 frames intervals from it. 
    This ensures a sufficient image overlap while gradually increasing the difficulty of the image pair, differentiating the break-down point of the competing alternatives. On SUN3D we use the same fifteen scenes and sampling procedure as~\cite{brachmann2019neural,moo2018learning,zhang2019oanet}. For both datasets we keep the original image resolution.

\begin{figure*}
    \centering
    \scriptsize
    \includegraphics[width=\textwidth]{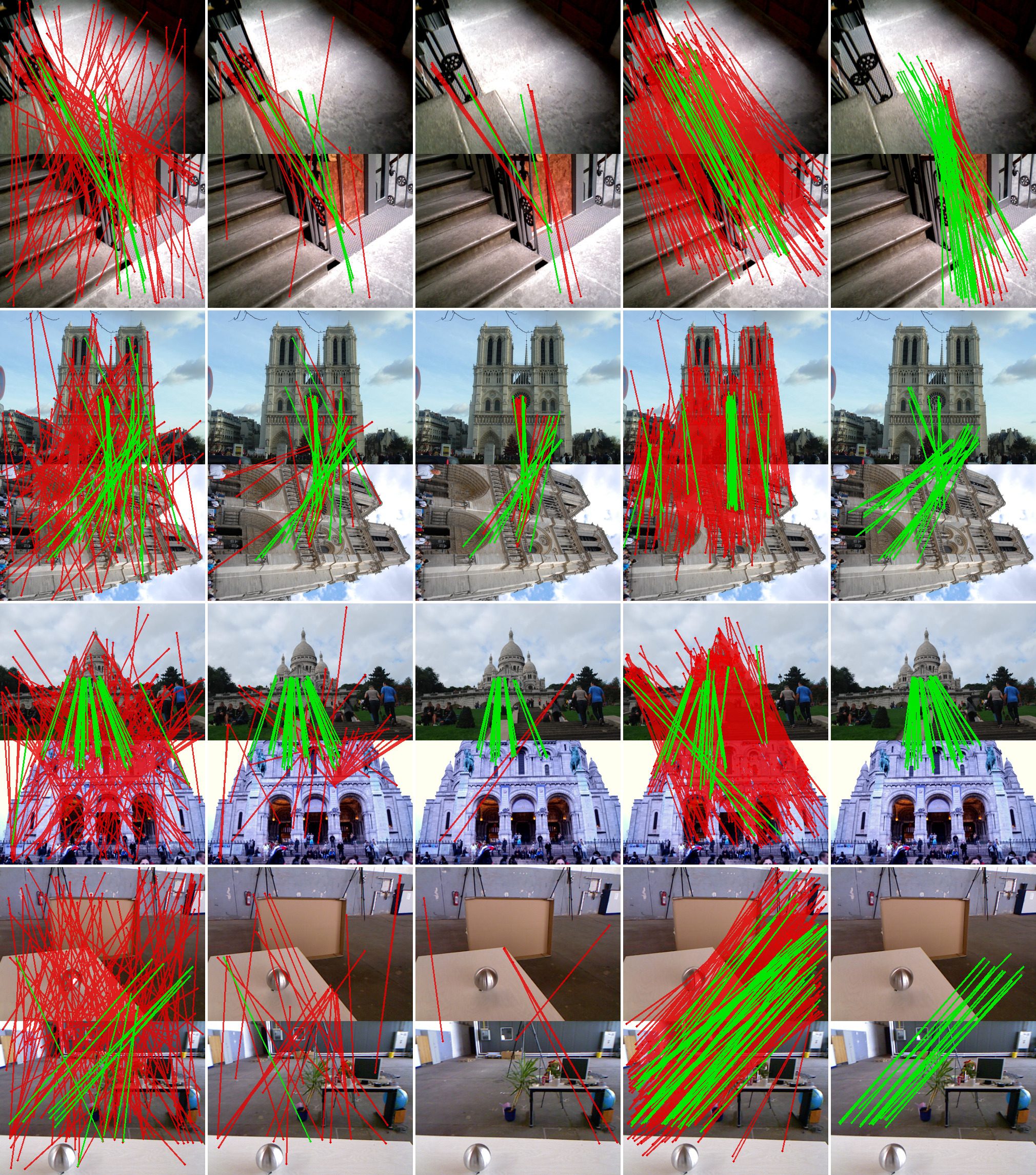}
    \newcommand{\cs}{0.175\textwidth}
    \begin{tabular}{C{\cs}C{\cs}C{\cs}C{\cs}C{\cs}}
        Ratio-test~\cite{lowe2004distinctive} & NGRANSAC~\cite{brachmann2019neural} & 
        GMS~\cite{bian2017gms} & OA-Net~\cite{zhang2019oanet} & Ours \\[-0.5em]
    \end{tabular}
    \caption{\textbf{Success cases} from our experiments. Matches agreeing with ground truth epipolar geometry are shown in green, others are in red. Examples include cases with very sparse correspondences, local repeated structures, weak texture, strong rotations and perspective deformations.}
    \label{fig:goodcases}
\end{figure*}
\begin{figure*}
    \centering
    \scriptsize
    \includegraphics[width=\textwidth]{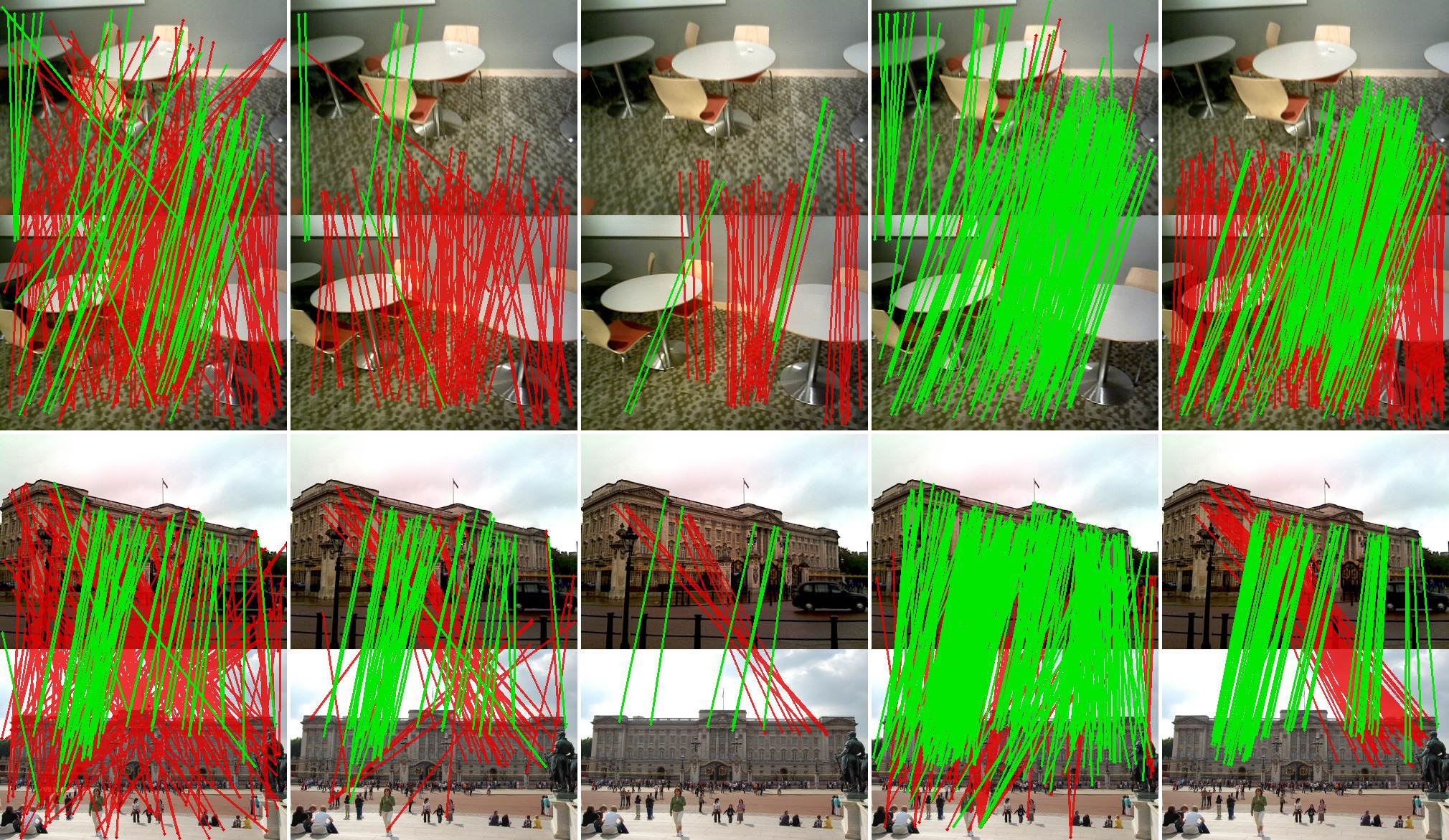}
    \newcommand{\cs}{0.175\textwidth}
    \begin{tabular}{C{\cs}C{\cs}C{\cs}C{\cs}C{\cs}}
        Ratio-test~\cite{lowe2004distinctive} & NGRANSAC~\cite{brachmann2019neural} & 
        GMS~\cite{bian2017gms} & OA-Net~\cite{zhang2019oanet} & Ours \\[-0.5em]
    \end{tabular}
    \caption{\textbf{Failure cases} from our experiments. Matches agreeing with ground truth epipolar geometry are shown in green, others are in red. The main failure case for our method is wide repeated structures along the image, which can locally mimic the correspondence distribution of the correct region.}
    \label{fig:badcases}
\end{figure*}

\subsection{Comparison with State of the Art} \label{subsec:compsota}
%
We compare our method against sample representatives of the current state of the art.
\begin{itemize}[itemsep=1pt,topsep=0pt,leftmargin=*]
    \item GMS (Grid-based Motion Statistics)~\cite{bian2017gms} is a non-learned method that models the statistics of having locally consistent matches and filters matches based on a statistical significance test over large groups. Designed with the objective of being fast, the authors use $10\,000$ ORB features~\cite{rublee2011orb}. 
    However, we found that with appropriate tuning the performances are higher using our SIFT setup with a ratio-test filtering beforehand, as suggested by the authors, thus we report these results using the public OpenCV implementation of the method with rotation and scale invariance.
    \item NGRANSAC (Neural Guided RANSAC)~\cite{brachmann2019neural} uses a neural network to predict sampling probabilities for RANSAC from keypoint location and ratio-test score. We use the pre-trained models provided by the authors for essential matrix estimation with SIFT keypoints pre-filtered with ratio-test of 0.8 (SIFT+Ratio+NG-RANSAC(+SI) label in~\cite{brachmann2019neural}), which have been trained on both YFCC100M~\cite{thomee2016yfcc100m} and SUN3D~\cite{xiao2013sun3d}. We experimentally found that, although the method outputs an essential matrix, better performance is achieved by using LO-RANSAC only on the inlier set found by NGRANSAC, thus after running both versions we report these results.
    \item OA-Net (Order Aware Network)~\cite{zhang2019oanet} learns to infer confidence scores on nearest neighbor matches looking at the global keypoint spatial consistency. They propose a soft assignment to latent clusters in canonical order, and an order-aware upsampling operation that restores the original size of the input to infer confidences. The authors provide a model pretrained on both YFCC100M~\cite{thomee2016yfcc100m} and SUN3D~\cite{xiao2013sun3d}. Our SIFT parameters are taken from the public implementation provided by the authors with the pretrained model. We tuned the default inlier threshold to optimize the Area Under Curve metric.
    \item We include a simple baseline using the standard ratio test with 0.8 threshold, as the default in SiftGPU~\cite{wu2011siftgpu} used in COLMAP~\cite{schoenberger2016sfm}. We also try the performance of this simple baseline with ten times more LO-RANSAC iterations, going from the $10^4$ used for all methods to $10^5$ iterations.
\end{itemize}
Table~\ref{tab:comparisons} reports the results of our experiments on both indoor and outdoor scenes. For comparability and deeper insights we report additional metrics in the supplementary material, including an upper bound approximation of the AUC used by some of the methods. All the competitor methods outperformed their original paper scores in our setup, where the main difference is the use of LO-RANSAC rather than OpenCV naive RANSAC. We found that local optimization can refine the solution by some degrees, improving the scores for low errors.

Results show that our method can drastically outperform current state of the art in both indoor and outdoor scenarios. While TUM is a completely new dataset for all learned methods, both OA-Net and NGRANSAC are trained on YFCC100M and SUN3D, although the scenes we evaluate on do not belong to their training set.

On Figures~\ref{fig:goodcases} and~\ref{fig:badcases} we report qualitative results that represent success cases and failure cases for our method with respect to the other baselines. Figure~\ref{fig:goodcases} shows how our method captures consistent global motion even when available correct matches are sparse, and is fully invariant to sharp rotation and scale changes. As affine coherence in keypoint patterns can give confidence to matches even when descriptors are ambiguous, our method is able to mine correspondences even from almost textureless surfaces or in the presence of locally repeating structures. However, this is not always the case for widely repeating regular structures, as illustrated in Figure~\ref{fig:badcases}. In such cases, there are one or more independent clusters of incorrect correspondences that locally exhibit significantly non-random patterns. Global approaches in this case have a chance to disambiguate the right cluster, and learned approaches can give priority to the cluster compatible with more likely motions, as OA-Net is doing.

\subsection{Ablation studies} \label{subsec:ablation}
We aim at understanding the contribution of each element we introduce in our method, thus we extensively evaluate different versions of our method subtracting one element at a time. For comparability with other methods, we run the same experiments in the same setting as in Section~\ref{subsec:compsota} on TUM and YFCC100M.
We target three optional steps in our pipeline and re-evaluate removing one or multiple of them:
\begin{enumerate}[itemsep=1pt,topsep=0pt,leftmargin=*]
    \item \textbf{AdaLAM}: the full method described in this paper
    \item \textbf{No-Side}: AdaLAM without side information filtering, \ie the condition in Eq.~\eqref{eq:corrinclusionsideinfo}.
    \item \textbf{No-Refit}: AdaLAM without refitting the estimated affinities on the set of inliers.
    \item \textbf{LAM}: locally-affine matching without adaptive thresholding in the affine fitting, but using a single fixed threshold. This is in sum a well-engineered baseline integrating several known good practices. We run this experiment with a wide range of thresholds and report only the best according to AUC5 and matching F1-score. Moreover, we observed consistently better performance of the LAM baseline without refitting, so we report these results.
\end{enumerate}
\begin{table}[tb]
\scriptsize
\centering
\caption{\textbf{Ablation tests} with varying setups of our method. The numbers are comparable with Table~\ref{tab:comparisons}. Areas under the curve (AUC) are in percentage. F1-score is computed with ground truth inliers. Times in milliseconds include nearest neighbor search and outlier rejection running on an RTX2080Ti GPU. Results and timings for OA-Net~\cite{zhang2019oanet} are additionally reported for better runtime comparability.}
\label{tab:ablation}
\setlength{\tabcolsep}{1.4mm}
\begin{tabular}{lcccccccc}
\toprule
 Method &  
 \multicolumn{4}{c}{TUM~\cite{sturm2012benchmark}} & 
 \multicolumn{4}{c}{YFCC100M~\cite{thomee2016yfcc100m}}\\
\cmidrule(r){2-5}
\cmidrule(l){6-9}

& AUC5 & AUC20 & F & time & AUC5 & AUC20 & F & time\\

\midrule
AdaLAM & \bf 27.4& \bf 53.4& \bf 49.1 & 14ms & \bf 58.8& \bf 82.4 & 59.4 & 22ms\\
No-Side     & 23.2& 48.8& 48.0 & 19ms & 57.3& 81.1 & 57.6 & 30ms\\
No-Refit    & 26.4& 51.4& 48.3 & 12ms & 58.6& 82.0 & 59.7 & 20ms\\
$LAM_{AUC}$    & 27.0 & 50.0 & 46.4 & \bf 11ms & 58.1 & 82.0 & 57.4 & 18ms\\
$LAM_{F}$    & 25.5 & 50.2 & 48.3 & \bf 11ms & 57.6 & 81.6 & 60.2 & 18ms\\
\midrule
OA-Net~\cite{zhang2019oanet}      & 22.1& 43.8& 48.3 & 21ms & 54.8 & 77.5 & \bf66.6& 37ms\\

\bottomrule
\end{tabular}
\end{table}
We report the results of our ablation in Table~\ref{tab:ablation}.
We measure a runtime of 15-25ms on image pairs with 4000-8000 extracted keypoints for AdaLAM, running on an RTX2080Ti. Since many of our baseline methods provide CPU implementations, or important CPU preprocessing steps, their runtimes are usually higher and not directly comparable. However, we found that the public implementation of OA-Net~\cite{zhang2019oanet} also performs all operations on PyTorch as we do. We measure runtimes of 20-40ms on the same hardware and keypoint collections. For comparison with the ablations, we also report the performance of OA-Net~\cite{zhang2019oanet} in Table~\ref{tab:ablation}.

We can observe that the LAM baseline, although not involving any novel component, is already more than competitive with state of the art methods in terms of Area Under Curve, and comparable in terms of F1-score. AdaLAM achieves comparable or superior results to both tuned baselines in both metrics.

As evident from the "No-Side" results, smart classical filters can increase both runtime and quality as they reduce the size of the problem by pruning grossly incorrect correspondences since the beginning.

\subsection{Aachen Day-Night Challenge} \label{subsec:aachendaynight}
The Aachen Day-Night dataset~\cite{sattler2018benchmarking,sattler2012image} allows to measure pose accuracy achieved when trying to localize nighttime query images against a 3D model built from daytime images. 
We follow the setup of the Local Feature Challenge from the CVPR 2019 workshop on ``Long-Term Visual Localization under Changing Conditions" and use the code provided by the organizers\footnote{\url{https://github.com/tsattler/visuallocalizationbenchmark/}}, but use our matching method rather than the default mutual nearest neighbor matching. 
Following~\cite{sattler2018benchmarking}, we report the percentage of nighttime queries localized within thresholds on their rotation and position errors with respect to the ground truth poses (three sets of thresholds are used: (0.5m, 2\degr) / (1m, 5\degr) / (5m, 10\degr)).

While currently the majority of top-scoring methods are using specialized learned local features, we use our method for outlier rejection on upright RootSIFT~\cite{lowe2004distinctive} features. In Table~\ref{tab:aachen} we report scores for the available baselines and with our method. We also run additional baselines using upright RootSIFT with simple filters such as mutual nearest neighbor and ratio-test for better comparability. 
We observe that our outlier rejection greatly improves localization performance on SIFT keypoints and descriptors, elevating it to comparable levels as the current state-of-the-art learned local features. 

\begin{table}[tb]
\scriptsize
\centering
\caption{\textbf{Aachen Day-Night:} percentage of nighttime queries localized within given accuracy measures of the ground truth poses}
\label{tab:aachen}
\setlength{\tabcolsep}{1.4mm}
\begin{tabular}{lccc}
\toprule

Method & (0.5m, 2\degr) & (1m, 5\degr) & (5m, 10\degr)\\
\midrule
UprightRootSIFT (public baseline) & 33.7 & 52.0 & 65.3 \\
UprightRootSIFT + Mutual Nearest Neighbor & 37.8 & 56.1 & 76.5	\\
UprightRootSIFT + Ratio-Test (0.8) & 41.8 & 57.1 & 75.5	\\
UprightRootSIFT + AdaLAM & \bf 45.9 &\bf  64.3 &\bf 86.7	\\
\midrule
UR2KID Scape Technologies\cite{yang2020ur2kid} & 46.9 & 67.3 & 88.8 \\
D2-Net - single-scale~\cite{dusmanu2019d2} & 45.9 & 68.4 & 88.8 \\
R2D2 V2 20K~\cite{revaud2019r2d2} & 46.9 & 66.3 & 88.8 \\
Dense-ContextDesc10k\_upright\_OANet~\cite{luo2019contextdesc,zhang2019oanet} & 48.0 & 63.3 & 88.8 \\
densecontextdesc10k\_upright\_mixedmatcher~\cite{luo2019contextdesc} &46.9 & 65.3 & 87.8\\
\bottomrule
\end{tabular}
\end{table}

\subsection{Image Matching Challenge} \label{subsec:imagematchingchallenge}
The Image Matching Challenge (CVPR 2020)~\cite{jin2020image} aims at comparing the performance of different methods in the task of stereo relative pose estimation and multiview reconstruction with off-the-shelf structure from motion software. Competing methods need to detect, describe and match keypoints in images. The challenge is divided into two tasks:
\begin{itemize}[itemsep=1pt,topsep=0pt,leftmargin=*]
    \item \textbf{Stereo}: the stereo task consists in establishing sparse correspondences between image pairs to estimate the fundamental matrix and consequently the relative pose of the image pair. Competing methods are ranked according to the mean average precision for a maximum pose error under 10 degrees.
    \item \textbf{Multiview}: the multiview task feeds the competing method's image correspondences to colmap~\cite{schoenberger2016sfm} to run structure from motion on multiple images. Methods are ranked according to the mean average precision for the maximum pose error under 10 degrees for all the reconstructed views.
\end{itemize}
The benchmark's test set includes several scenes from YFCC100M, however there is no overlap with the subset used in our previous experiments.

Our submission to the challenge uses classical DoG keypoint detection with HardNet~\cite{mishchuk2017working} upright descriptors out of the box as provided by the challenge organizers, extracting at most 8000 keypoints per image. Nearest neighbor matches are filtered using AdaLAM with no hyperparameter tuning, and subsequently filtered with 100k iterations with DEGENSAC~\cite{frahm2006ransac}. The final pose is obtained by estimating the fundamental matrix with the 8-point algorithm. We report results for our submission and other baselines in Table~\ref{tab:imagematchingchallenge}. The HardNet-Upright + RT + DEGENSAC baseline has been provided by the challenge organizers, using the same HardNet and DEGENSAC setup as ours, and optimizing the ratio-test threshold separately for each task on a small validation set. AdaLAM outperforms the ratio-test baseline on both tasks without requiring any hyperparameter tuning. We also report the other three top performing methods from the challenge for comparison.
\begin{table}[tb]
\scriptsize
\centering
\caption{\textbf{Image Matching Challenge:} Results for the Image Matching Challenge 2020. We report mAP10 for the stereo task (ST), multiview (MV) and overall score (All). The methods marked with an asterisk use different descriptors than ours.}
\label{tab:imagematchingchallenge}
\setlength{\tabcolsep}{1.4mm}
\begin{tabular}{lccc}
\toprule

Method & ST & MV & All\\
\midrule
HardNet-Upright + AdaLAM + DEGSAC & 58.3 & 77.1 & 67.7 \\
HardNet-Upright + RT + DEGSAC & 57.3 & 72.3 & 64.8 \\
\midrule
Guided-Hardnet*& \bf 61.1 & 78.3 & \bf 69.7 \\
Guided-Hardnet* + OANet & 60.3 & \bf 78.6 & 69.4 \\
ContextDesc-Upright* + MNN + OANetV2.1 + DEGSAC & 57.8 & 77.0 & 67.4\\
\bottomrule
\end{tabular}
\end{table}

\section{Conclusions}
In this paper we proposed AdaLAM, a method to reject outliers from a set of putative correspondences, inspired by local consistency constraints which have been re-discovered repeatedly in the last years~\cite{lowe2004distinctive,zhang1995robust,jung2001robust,jegou2008hamming,sattler2009scramsac,bian2017gms,lin2017code,lin2016repmatch}. We formulated our approach as a highly parallel algorithm to be run on modern GPUs in the order of the tens of milliseconds for several thousand keypoints per image. We show that, with the proposed adaptive relaxation of the underlying assumptions for local consistency, we can improve over simple affine consistency filters, which are already very competitive. Our method can greatly outperform the current state of the art both in favorable settings, where the affine model can be more discriminative, and on unfavorable, less structured settings.


\appendix 

\begin{table*}
\centering
\caption{\textbf{RANSAC Comparison}: we compare OpenCV RANSAC and LORANSAC on ratio-test filtered matches to further confirm the superiority of LORANSAC in our pipeline.}
\label{tab:ransacs}
\setlength{\tabcolsep}{1.0mm}
\begin{tabular}{lccccccccc}
\toprule
 Method & 
 \multicolumn{3}{c}{TUM~\cite{sturm2012benchmark}} & 
 \multicolumn{3}{c}{SUN3D~\cite{xiao2013sun3d}} & 
 \multicolumn{3}{c}{YFCC100M~\cite{thomee2016yfcc100m}}\\
\cmidrule(r){2-4}
\cmidrule(rl){5-7}
\cmidrule(r){8-10}

  & AUC5 & AUC10 & AUC20 & AUC5 & AUC10 & AUC20 & AUC5 & AUC10 & AUC20\\
\midrule
RT + LO-RANSAC~\cite{chum2003locally}                      &\bf 16.1&\bf 24.8&\bf 33.6&\bf 5.9 &\bf 14.1&\bf 25.6 &\bf 51.9&\bf 64.9&\bf 76.3\\
RT + OpenCV RANSAC                                         & 9.2 & 16.3& 23.7& 2.1 & 5.2 & 10.2 & 19.4& 31.2& 43.1\\

\bottomrule
\end{tabular}
\end{table*}

\begin{table*}
\centering
\caption{\textbf{Extended metrics results}: for better comparability with previous and future works we report additional metrics of our results. Starred Area under the Curve is computed as the area below an histogram with five-degrees bins.}
\label{tab:extresults}
\setlength{\tabcolsep}{1.4mm}
\begin{tabular}{lcccccccccc}
\toprule
 Method &  
 \multicolumn{9}{c}{TUM~\cite{sturm2012benchmark}}\\
\cmidrule(rl){2-10}

 & AUC5 & AUC10 & AUC20 & AUC5* & AUC10* & AUC20* & mAP5 & mAP10 & mAP20\\
\midrule
AdaLAM &\bf 27.4&\bf 41.3& \bf53.4&\bf48.3&\bf 44.0&\bf  60.9&\bf 48.3&\bf  61.0& \bf 68.4\\
OA-Net & 22.1& 33.3& 43.7&  38.1&  43.7& 49.8&  38.1& 49.3& 57.3\\
NGR & 19.4& 29.6& 38.7&  33.4&  38.9& 44.1&  33.4& 44.4& 50.6\\
GMS& 19.6& 30.5& 41.3& 34.4 &  40.2& 47.0&34.4 &  46.0& 55.1\\
RT (10k) & 16.1& 24.8& 33.6&  27.7&  32.8& 38.5&  27.7& 38.0& 46.0\\
RT (100k)& 17.3& 26.6& 36.2&  29.3&  34.6& 41.1&  29.3& 39.9& 49.1\\
\midrule
&  
 \multicolumn{9}{c}{YFCC100M~\cite{thomee2016yfcc100m}}\\
\cmidrule(rl){2-10}
AdaLAM &\bf 58.8&\bf 72.0&\bf 82.4&\bf  78.4&\bf  83.8&\bf 88.8&\bf 78.4&\bf 89.2&\bf 94.6\\
OA-Net & 54.8& 67.3& 77.5&  71.6&  77.8& 83.2&  71.6& 84.1& 89.6\\
NGR & 53.8&  66.7&  77.7& 71.4&  78.1& 84.0&71.4 & 84.7 & 90.9\\
GMS&  52.3& 65.0& 76.0&  69.5&  76.2& 82.2&  69.5& 83.0& 89.1\\
RT (10k)&  51.9& 64.9& 76.3&  69.5&  76.4& 82.7&  69.5& 83.4& 89.9\\
RT (100k)&  53.2& 66.3& 77.5&  71.1&  78.0& 84.0&  71.1& 84.9& 90.9\\
\midrule
&  
 \multicolumn{9}{c}{SUN3D~\cite{xiao2013sun3d}}\\
\cmidrule(rl){2-10}
AdaLAM &\bf7.9 &\bf 19.2 &\bf 34.6 &\bf 19.9&\bf 29.4&\bf 42.0&\bf 19.9&\bf 39.0&\bf 58.4\\
OA-Net &7.0 & 16.5 & 29.7 &  17.1&  25.3& 36.0&  17.1& 33.5& 49.9\\
NGR & 6.2 & 15.0 & 27.3 &  15.5&  23.1& 33.2&  15.5& 30.6& 46.5\\
GMS &  6.8 & 15.9 & 29.1 &  16.6&  24.5& 35.4&  16.6& 32.3& 50.0\\
RT (10k)&  5.9 & 14.1 & 25.6 &  14.9&  21,7& 31.1&  14.9& 28.5& 43.5\\
RT (100k)& 6.1 & 14.5 & 26.3&  15.2&  22.4& 32.0&  15.2& 29.5& 44.6\\
\bottomrule
\end{tabular}
\end{table*}

\begin{table*}
\centering
\caption{\textbf{Tuned NGRANSAC baseline setup}: we run experiments on YFCC100M to tune the setup for NGRANSAC within our evaluation pipeline. We report the metrics explained in Section~\ref{sec:addm}.}
\label{tab:ngr}
\setlength{\tabcolsep}{1.4mm}
\begin{tabular}{lcccccccccc}
\toprule
 Method &  
 \multicolumn{9}{c}{YFCC100M~\cite{thomee2016yfcc100m}}\\
\cmidrule(rl){2-10}

 & AUC5 & AUC10 & AUC20 & AUC5* & AUC10* & AUC20* & mAP5 & mAP10 & mAP20\\
\midrule
Original & 39.4& 50.8& 59.5& 55.3& 60.6& 64.8& 55.3& 65.9& 69.7 \\
Refitting-ST & 53.2& 66.2& 77.2& \bf 71.5&  77.7& 83.7& \bf 71.5& 83.8& 90.5\\
Refitting-LT & \bf 53.8& \bf 66.7& \bf 77.7& 71.4&\bf  78.1& \bf 84.0&71.4 &\bf 84.7 &\bf  90.9\\

\bottomrule
\end{tabular}
\end{table*}

\section{RANSAC implementation choice}
\label{sec:rans}

In our evaluation pipeline we opted for an advanced RANSAC implementation, using the LO-RANSAC~\cite{chum2003locally} implementation provided by COLMAP~\cite{schoenberger2016sfm}, rather than a standard implementation as the one found in OpenCV. The superiority in pose estimation performance of recent state-of-the-art RANSACs~\cite{barath2019magsac,chum2003locally,raguram2012usac,torr2002napsac,ni2009groupsac} compared to the first proposed algorithm~\cite{fischler1981random} has been confirmed several times by each newly published method, by recent benchmarks~\cite{jin2020image}, and tested again for our specific pipeline as shown in Table~\ref{tab:ransacs}. This is not surprising as each new method builds on the initial idea from~\cite{fischler1981random}.

\section{Additional Metrics}
\label{sec:addm}

Table~\ref{tab:extresults} reports additional metrics for the experiments presented in the main paper for better comparability with previous and future work. In particular we add the two following metrics:

\begin{enumerate}
    \item \textbf{mAPX}: mean average precision under X degrees, i.e.~the rate (in percentages) of successful pose estimation considering as success a pose with maximum error lower than X degrees. This allows comparability of our results with the original OA-Net~\cite{zhang2019oanet} paper.
    \item \textbf{AUCX*}: approximate Area Under the Curve below X degrees. As some previous works~\cite{brachmann2019neural,moo2018learning} report AUC measures approximated as the cumulative area under a histogram with 5-degrees bins, we report the same for comparability with them.
\end{enumerate}

The same observations drawn from the exact AUC are drawn from the extended metrics: our method greatly outperforms current state of the art in both indoor and outdoor scenarios.

\section{Neural Guided RANSAC Baseline}
\label{sec:ngr}

We observed that Neural Guided RANSAC (NGRANSAC)~\cite{brachmann2019neural} performs better by re-fitting the essential matrix with LO-RANSAC~\cite{chum2003locally} on its inlier set rather than directly using the essential matrix that is the output of the authors code. In Table~\ref{tab:ngr}, we report comparative results of using NGRANSAC as originally designed and with re-fitting. We test on three settings:

\begin{enumerate}
    \item \textbf{Original}: this is the original NGRANSAC setup as suggested by the authors, reproducing the results in~\cite{brachmann2019neural} for the SIFT+Ratio+NG-RANSAC (+SI) label on essential matrix estimation. As default in the sample code from the authors, we use a RANSAC threshold of 0.001 and run 25000 iterations. We then evaluate directly the essential matrix found by NGRANSAC.
    \item \textbf{Refitting-ST}: in this alternative we run the whole NGRANSAC pipeline exactly as in the \textit{Original} case, with the same setup and threshold, and then use the inliers found to fit the essential matrix using the same LO-RANSAC setup used by all other methods. We found that the default threshold of 0.001, suggested originally for the method, is too strict and reduces the space for local optimization inside LO-RANSAC as the chosen inliers are already strictly agreeing on some model. Therefore, we found that larger thresholds were better for this setup.
    \item \textbf{Refitting-LT}: this is the setup that we used and ran on all experiments reported in the main paper. We run the \textit{Original} pipeline on essential matrix estimation with 25000 iterations and a threshold of 0.01, and then fit the final essential matrix with LO-RANSAC on the inlier set found by NG-RANSAC. A higher threshold than the original allows more space for the local optimization to find better poses on average, as seen in Table~\ref{tab:ngr}.
\end{enumerate}

{\small
\bibliographystyle{ieee_fullname}
\bibliography{egbib}
}

\end{document}